\documentclass{article}

\usepackage[preprint]{neurips_2019}

\usepackage[utf8]{inputenc} 
\usepackage[T1]{fontenc}    
\usepackage{url}            
\usepackage{booktabs}       
\usepackage{amsfonts}       
\usepackage{nicefrac}       
\usepackage{microtype}      
\usepackage{xfrac}

\usepackage{multirow}       

\usepackage{mathtools}
\usepackage{natbib}
\usepackage{xcolor}
\usepackage{url}
\usepackage{amsmath}
\usepackage{amssymb}
\usepackage{graphicx}
\usepackage{subcaption}
\usepackage{nicefrac}
\usepackage{appendix}
\usepackage{enumitem}
\definecolor{darker}{rgb}{0,0.15,0.7}
\usepackage[colorlinks, urlcolor=darker, citecolor=darker, linkcolor=darker]{hyperref} 
\setcitestyle{round}


\usepackage{algorithm}
\usepackage{algorithmic}

\usepackage{amsthm}

\begingroup
    \makeatletter
    \@for\theoremstyle:=definition,remark,plain\do{%
        \expandafter\g@addto@macro\csname th@\theoremstyle\endcsname{%
            \addtolength\thm@preskip\parskip
            }%
        }
\endgroup

\title{Multi-Speaker End-to-End Speech Synthesis}
\author{
Jihyun Park\thanks{These authors contributed equally to this work. Correspondence to $<$zhaokexin01@baidu.com$>$.}~\ \thanks{Work done while interning at Baidu Research.} 
\quad \ Kexin Zhao$^{\ast}$$^{\ddagger}$ 
\quad \ Kainan Peng$^{\ddagger}$ \quad \ Wei Ping$^{\ddagger}$
\vspace{0.15em} \\
\texttt{jihyunp@ics.uci.edu, \{zhaokexin01,pengkainan,pingwei01\}@baidu.com}
\vspace{0.3em}
\\
$^{\dagger}$University of California, Irvine, Department of Computer Science, Irvine, CA
\\
$^{\ddagger}$Baidu Research, Sunnyvale, CA}
\begin{document}

\maketitle

\begin{abstract}
In this work, we extend ClariNet~\citep{Ping2018}, a fully end-to-end speech synthesis model (i.e., text-to-wave), to generate high-fidelity speech from multiple speakers. 
To model the unique characteristic of different voices, low dimensional trainable speaker embeddings are shared across each component of ClariNet and trained together with the rest of the model. 
We demonstrate that the multi-speaker ClariNet outperforms state-of-the-art systems in terms of naturalness, because the whole model is jointly optimized in an end-to-end manner.~\footnote{Synthesized speech samples can be found in: \url{https://multi-speaker-clarinet-demo.github.io/} .}
\end{abstract}

\section{Introduction}
\label{sec:intro}
There have been continuous efforts on synthesizing high-fidelity speech by computer. These so-called text-to-speech (TTS) systems have various applications in human-computer interaction, assistive technology, media and entertainment. 
In particular, deep learning-based TTS systems have evolved at a rapid pace, starting from having multiple finely engineered neural networks for different components~(e.g., phoneme duration, fundamental frequency predictions)~\citep{ze2013statistical, dv1, dv2}, to text-to-spectrogram models connected with a separate vocoder~\citep{tacotron, dv3, char2wav}.
ClariNet~\citep{Ping2018} is the first text-to-wave neural architecture for speech synthesis, which converts text to raw waveform using a single neural network trained from scratch. 
However, the original ClariNet is a single speaker system and not aimed for generating various voices from different speakers.

In this paper, we augment ClariNet by adding the multi-speaker functionality to the system. Applying a similar technique as Deep Voice 3~\citep{dv3},  we add speaker embedding as a bias to each part of the network.
We demonstrate that this modified version of ClariNet can generate unique voices of more than one hundred speakers in higher quality than the state-of-the-art methods.


\section{Related work}
Deep learning-based TTS systems~\citep{ling2015deep} can be categorized according to the structure of the systems.
%
Earlier systems including Deep Voice 1~\citep{dv1} and Deep Voice 2~\citep{dv2} retain the traditional parametric TTS pipeline, which has separate grapheme-to-phoneme, segmentation, phoneme duration, frequency, and waveform synthesis models. Building and tuning such system require laborious feature engineering. In addition, each component within the system is optimized separately and their errors can accumulate through the pipeline, which may lead to sub-optimal performance, e.g., unnatural prosody.

Deep Voice 3~\citep{dv3}, Tacotron~\citep{tacotron}, Tacotron 2~\citep{tacotron2}, Char2Wav~\citep{char2wav}, and ParaNet~\citep{paranet} are attention-based seq2seq models, which convert text to acoustic features~(e.g., mel spectrogram) and have much more compact architectures. 
To synthesize high-fidelity speech, one still needs to feed the predicted spectrogram to a separately trained neural vocoder ~\citep{dv3, tacotron2}, such as WaveNet~\citep{wavenet}. 
As demonstrated in \citep{Ping2018}, it may still result in sub-optimal performance as text-to-spectrogram model and WaveNet are optimized separately.
In contrast, ClariNet~\citep{Ping2018} uses a variant of WaveNet conditioned on the hidden representation within the model, which is jointly trained with the whole model in an end-to-end manner. 

Building multi-speaker TTS systems is a long-standing task in speech synthesis research.
In traditional parametric TTS system~\citep{yamagishi2009robust}, an average voice model is trained using all speakers' data, which is then adapted to different speakers. 
There have been many recent works on neural network-based multi-speaker TTS models in addition to Deep Voice 2 and Deep Voice 3. 
Several deep learning-based systems~\citep{fan2015multi,yang2016training} also build average voice model, with i-vectors as inputs and speaker-dependent output layers for different speakers. 
Wu et al. \citep{wu2015study} empirically studies and compares DNN-based multi-speaker modeling methods. 
VoiceLoop~\citep{voiceloop} focuses on training the model with in-the-wild dataset and shows that a simple buffer structure could produce high quality voices.
All of these systems are built on the pipelines with separately trained components.
Our model is the first multi-speaker text-to-wave system that is trained end-to-end from scratch and can produce high quality synthesized speech for a hundred speakers.

\section{Multi-speaker ClariNet}

\begin{figure*}[tb] \centering
\hspace{-.3cm}
\includegraphics[clip, width=0.7\textwidth]{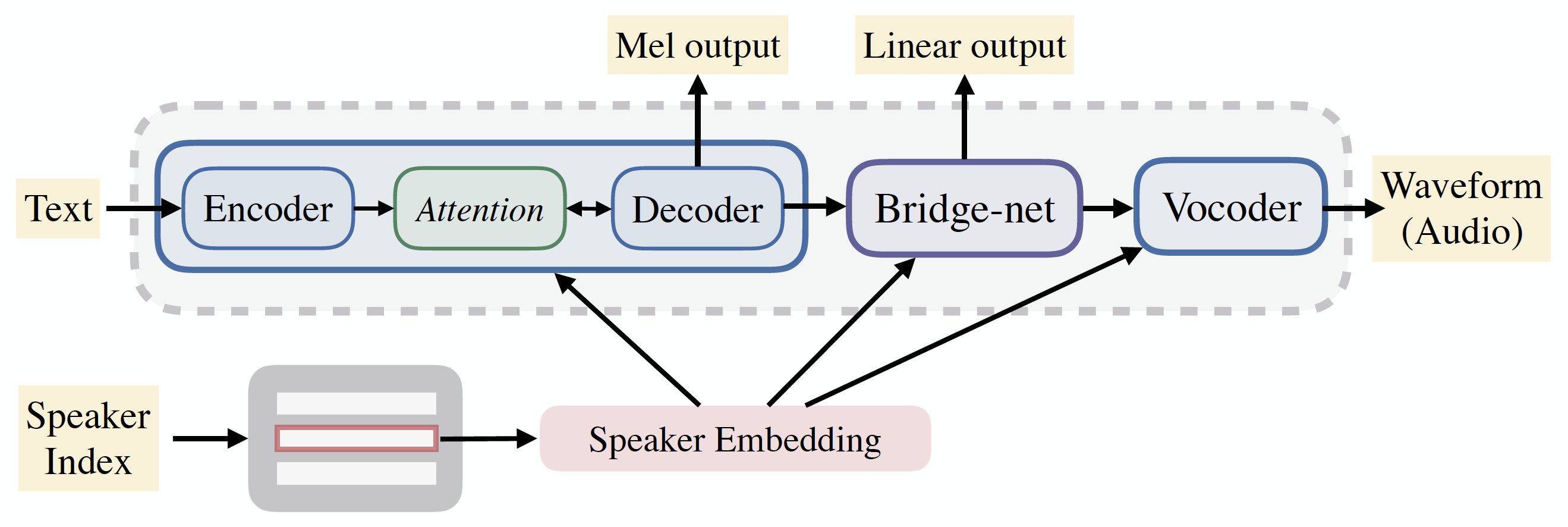}
\caption{Overall architecture of multi-speaker ClariNet.}
\label{fig:overall_arc}
\end{figure*}
\vspace{0.1em}

\begin{figure*}[tb] \centering
\hspace{-.3cm}
\includegraphics[clip, width=0.78\textwidth]{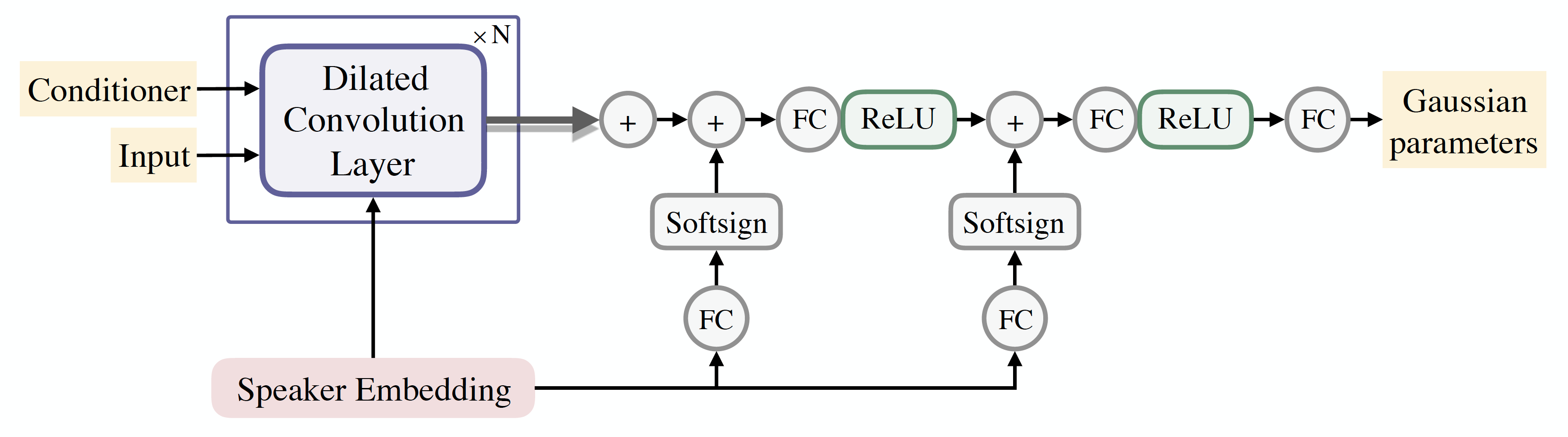}
\caption{Overall architecture of the Gaussian autoregressive WaveNet. Shadowed arrow after the dilated convolution layers depicts the skip connections, and GAU stands for gated activation unit.}
\label{fig:wavenet}
\end{figure*}

\begin{figure*}[tb] \centering
\includegraphics[clip, width=0.37\textwidth]{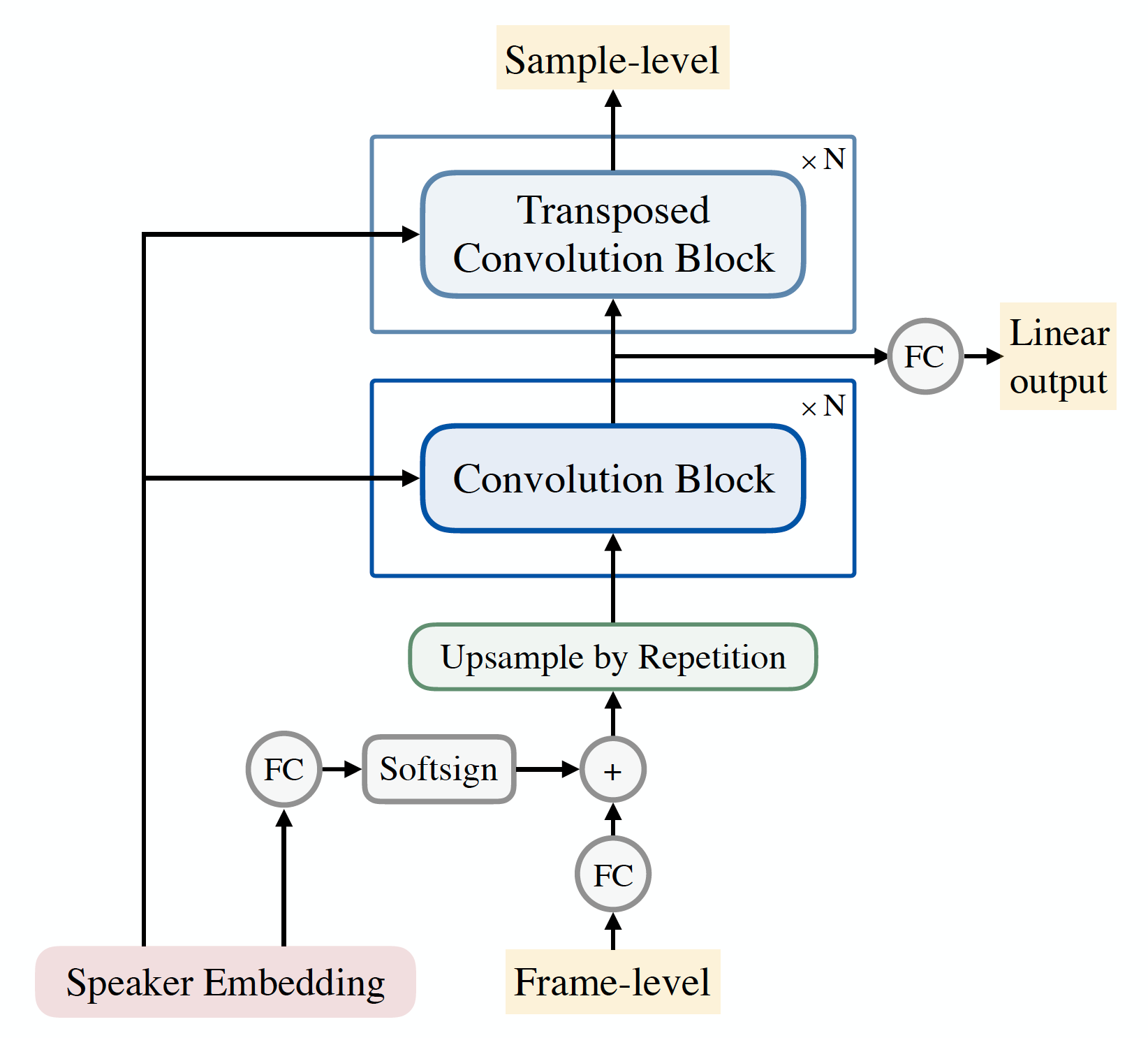}
\hspace{10pt}
\includegraphics[clip, width=0.28\textwidth]{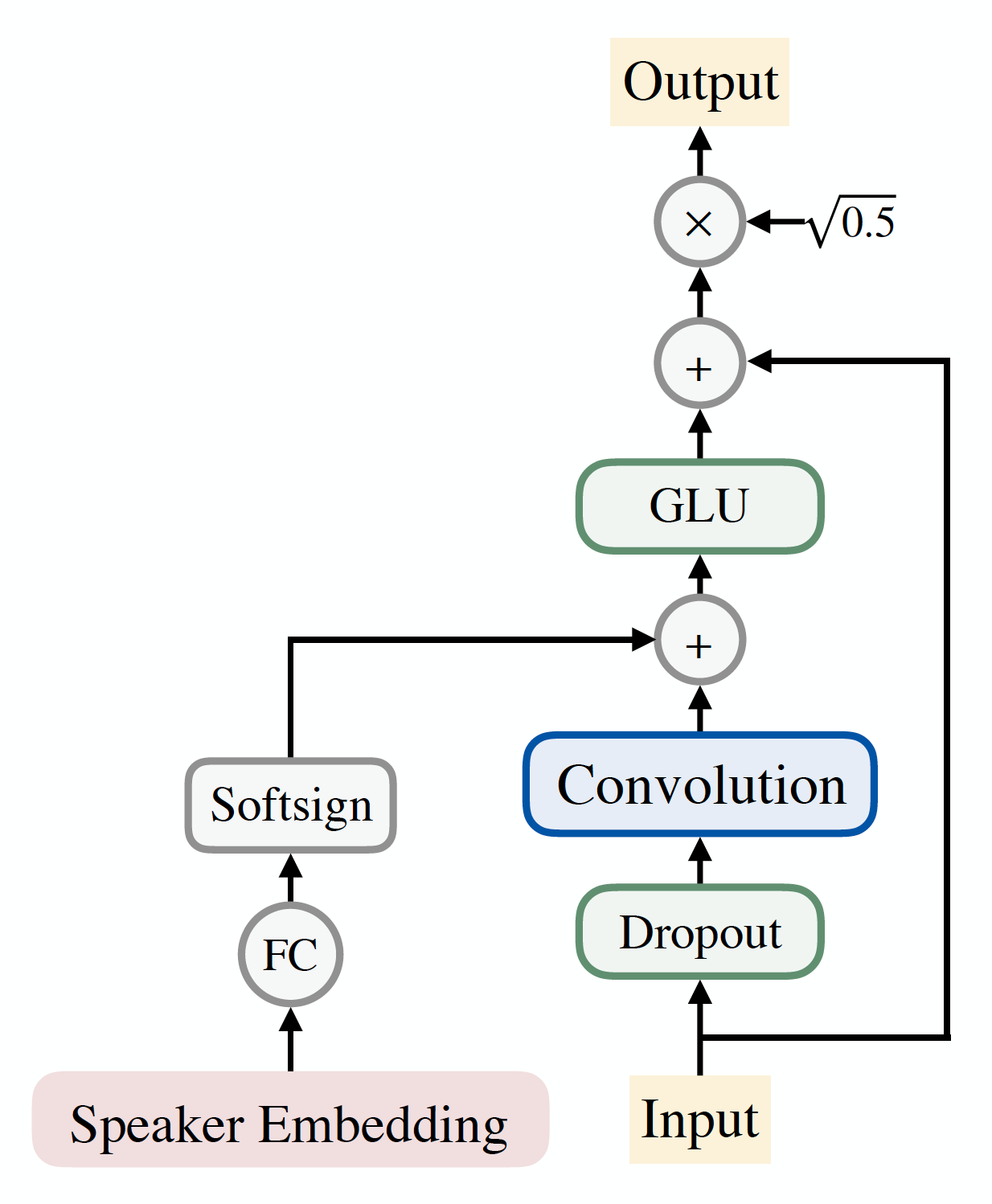}\\
\hspace{25pt} (a) \hspace{135pt} (b) \\
\includegraphics[clip, width=0.3\textwidth]{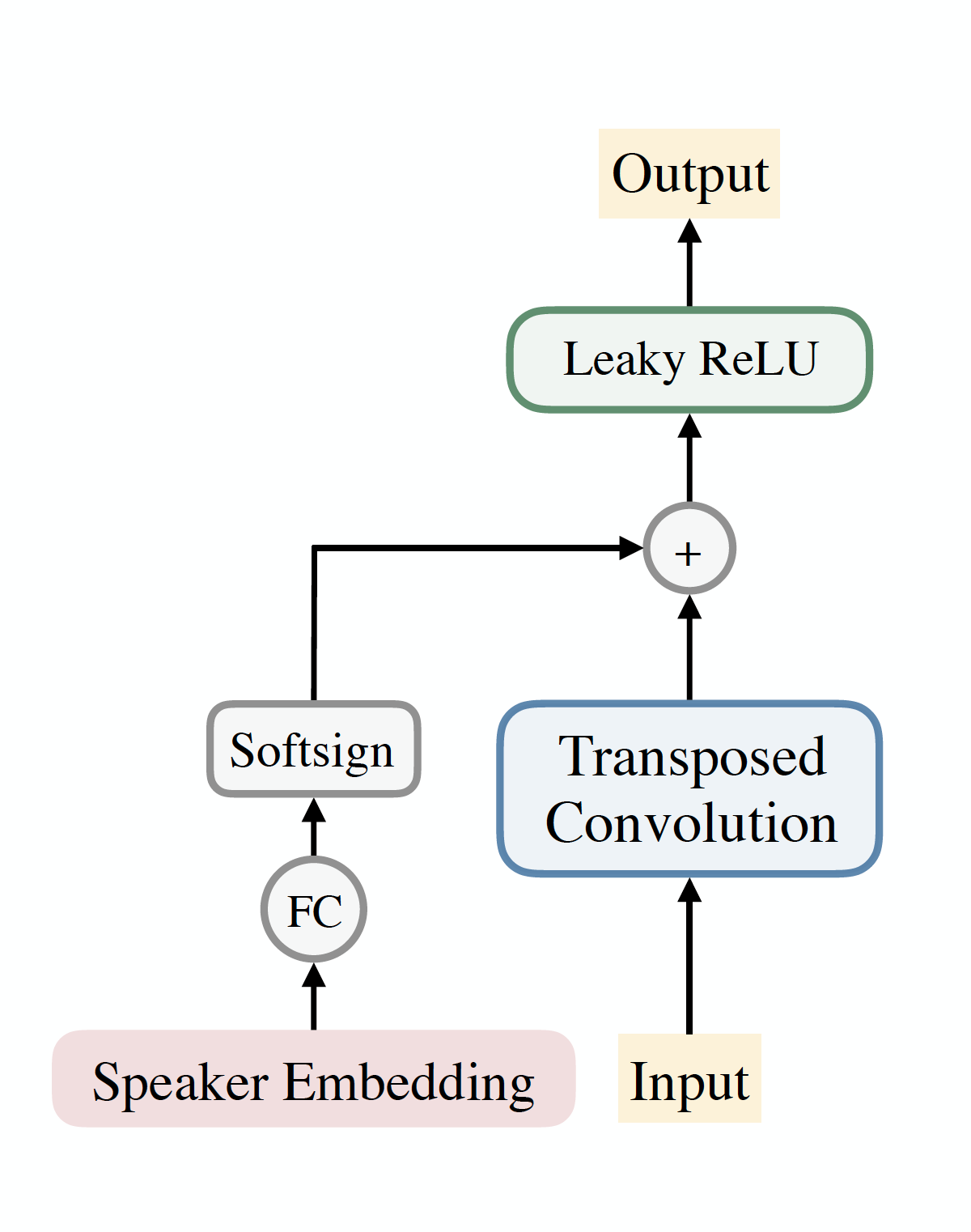}
\hspace{40pt}
\includegraphics[clip, width=0.28\textwidth]{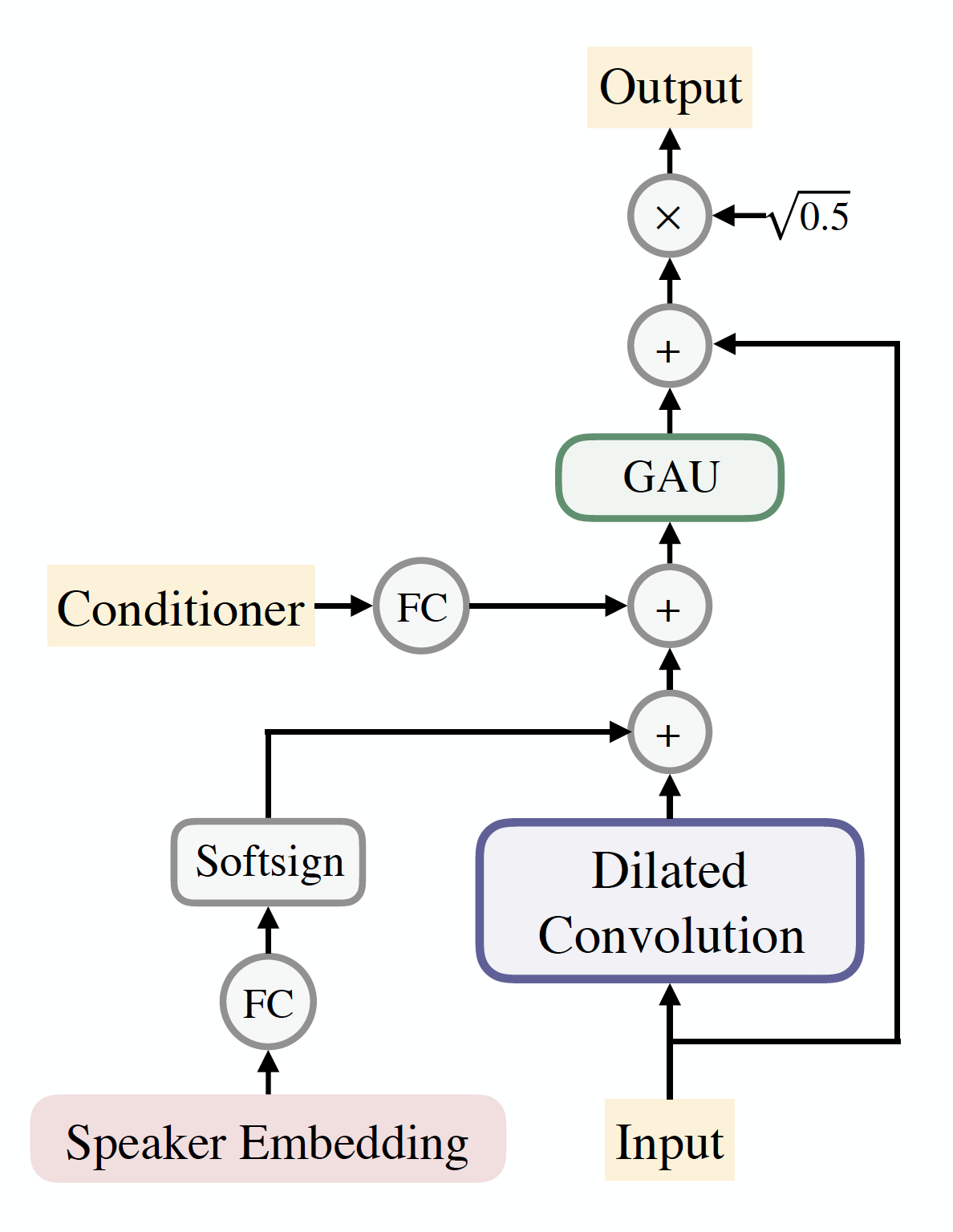}\\
\hspace{25pt} (c) \hspace{135pt} (d) \vspace{0pt}
\caption{Architectures of (a) Bridge-net, (b) Convolution block, (c) Transposed convolution block, and (d) Dilated convolution layer.}
\label{fig:convblocks}
\end{figure*}

\subsection{Architecture}
We follow a similar architecture as the original ClariNet, which trains the text-to-wave model in an end-to-end manner. Figure~\ref{fig:overall_arc} depicts the overall architecture of the multi-speaker ClariNet. Instead of connecting two separately trained models, one for generating spectral features~(e.g., mel spectrogram) and the other for synthesizing raw waveform, the model feeds the hidden state from the first component to the vocoder through the \emph{bridge-net} and trains the whole model jointly.

In addition, trainable speaker embeddings serve as conditional input throughout the whole network and are stored in an embedding lookup table. The projected embeddings after a softsign function are added as a bias to the convolution blocks within the encoder and decoder networks as in \citep{dv3}. They are also added to the bridge-net and the vocoder in a similar manner. Since ClariNet is an end-to-end network where everything is jointly learned, adding speaker embeddings to each component of the model allows them to learn better representations of the speaker characteristics, which in turn improves the quality of the generated audios.

Single-speaker ClariNet consists of four components including encoder, decoder, bridge-net, and vocoder. Since the same encoder-decoder structure as in Deep Voice 3~\citep{dv3} and ClariNet~\citep{Ping2018} is used for multi-speaker ClariNet, we skip the detailed description of the encoder and decoder due to the space constraint and focus on the architecture of the \textit{speaker-dependent} bridge-net and WaveNet.


\subsubsection{Bridge-net}
Bridge-net is a non-causal convolutional processing block that connects the two main parts of the system. It consists of a series of convolution blocks followed by a series of transposed convolution blocks. The speaker embedding is added as a bias to each block after being projected by a fully connected layer and softsign activated. The overall architecture of bridge-net is shown in Figure \ref{fig:convblocks} (a).
The non-causality of the convolution blocks allows utilizing future information, which gathers more knowledge for prediction. Also, transposed convolution is used to upsample the hidden representation obtained from the convolution blocks from frame-level to sample-level. We can see the diagrams for the convolution block and the transposed convolution block in Figures \ref{fig:convblocks} (b) and (c), respectively.


\subsubsection{Vocoder}
As in single-speaker ClariNet, a Gaussian autoregressive WaveNet is used as the vocoder (shown in Figure \ref{fig:wavenet}). WaveNet consists of multiple dilated convolution layers and the speaker embedding is fed to each layer as a bias. Figure \ref{fig:convblocks} (d) shows the detailed diagram of a single dilated convolution layer. The speaker embedding is also added to the input of each fully-connected layer at the end of the architecture after a series of dilated convolution layers. Note that the transposed convolution blocks in bridge-net upsample the frame-level hidden representations to sample-level and feed these information as the local conditioner input to the vocoder. In this paper, we only discuss the autoregressive WaveNet without knowledge distillation. 

\subsubsection{Objective function}
We optimize a linear combination of the losses from decoder, bridge-net and vocoder. In our experiments, we set the linear coefficients to 1 for simplicity. 
Instead of predicting the end of the utterance with the \emph{done} loss  as in Deep Voice 3, we simply keep track of the most recent value of a monotonic attention layer and stop when it reaches the last character, which is the period (``.'').

\section{Experiments}
In this section, we present several experiments to evaluate the proposed multi-speaker ClariNet.~\footnote{Demo website: \url{https://multi-speaker-clarinet-demo.github.io/}}

\subsection{Data and preprocessing}
We train our multi-speaker ClariNet model on the VCTK dataset \citep{vctk2009}, which consists of audio recordings from 108 speakers, with approximately 400 utterances for each speaker and a total duration of $\sim$44 hours. The audios are downsampled from 48 KHz to 24 KHz. Also, we clip the leading silence of audios but keep the trailing silence. Keeping the trailing silence helps preventing the clicking sound or sharp noise at the end of the synthesized audios. We use log-mel spectrogram with 80 bands as decoder input. 
·
\subsection{Multi-speaker ClariNet} 
Similar to \citep{dv1}, the WaveNet component of our multi-speaker ClariNet model employs a stack of dilated convolution blocks, where each block has 10 layers and the dilation is doubled at each layer, $\{1,2,4,\ldots,512\}.$ We add the output hidden states of 128 channels from each layer through skip connections. We use a single Gaussian as the output distribution for the WaveNet vocoder and $-7$ as the lower bound for natural logarithm of the predicted Gaussian standard deviation \citep{Ping2018}.
We use 8 GPUs to train three multi-speaker ClariNet models with 20, 30, and 40 dilated convolution layers in its WaveNet component, respectively.
All three models are trained 
for 1.5M steps  
using Adam optimizer \citep{Kingma2015} with batch size $16$. The learning rate is initially set to 0.001, and it gets annealed by half at every 200K steps after the first 500K steps. Additional hyperparameter settings are listed in Table \ref{tab:params}.

\begin{table}[tb]
    \caption{Shared hyperparameters used for the three multi-speaker ClariNet models in our experiment.}
    \vspace{0.1em}
    \centering
    \begin{tabular}{l c}
    \hline
        Parameter & Value \\\hline
        FFT Size & 2048\\
        FFT Window Size / Shift &  1200 / 300\\
        Reduction Factor & 4\\
        Speaker Embedding Dim. & 32\\
        Character Embedding Dim. & 256\\
        Encoder Layers / Conv. Width / Channels & 7 / 5 / 128\\
        Decoder Pre-net Affine Channels & 128, 256\\
        Decoder Layers / Conv. Width & 6 / 5\\
        Attention Channels & 256\\
        Positional Encoding Weight / Initial Rate & 0.1 / 7.6\\
        Bridge-net Conv Layers / Width / Channels & 6 / 5 / 256\\
        Dropout Keep Probability & 0.95\\
        Max. Gradient Norm & 100.0\\
        Gradient Clipping Max. Value & 5.0\\\hline
    \end{tabular}
    \label{tab:params}
\end{table}


\begin{figure*}[tb]
    \centering
    \includegraphics[width=0.4\textwidth]{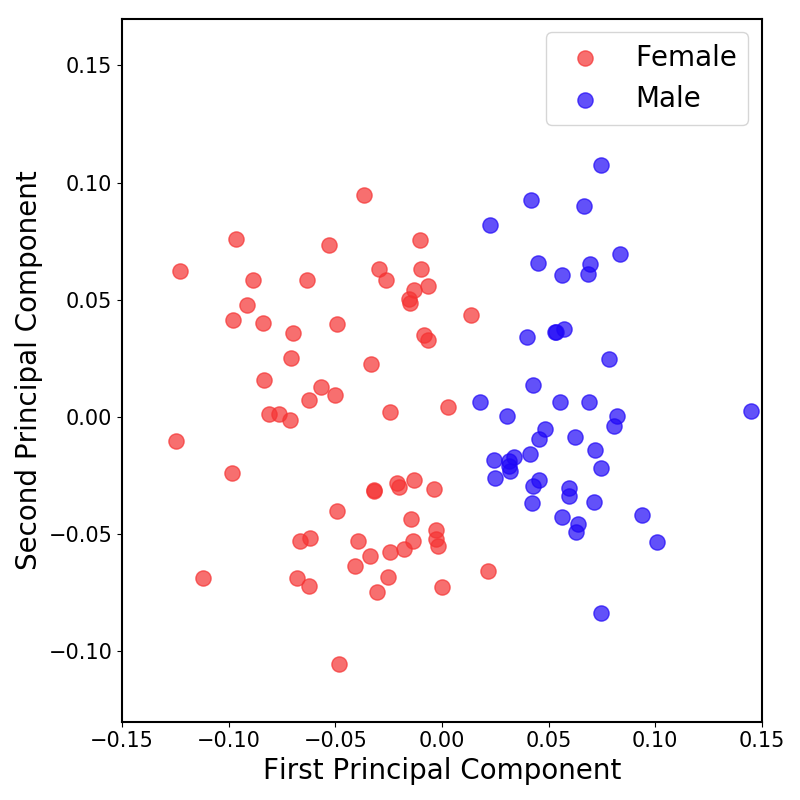} \hspace{35pt}
    \includegraphics[width=0.4\textwidth]{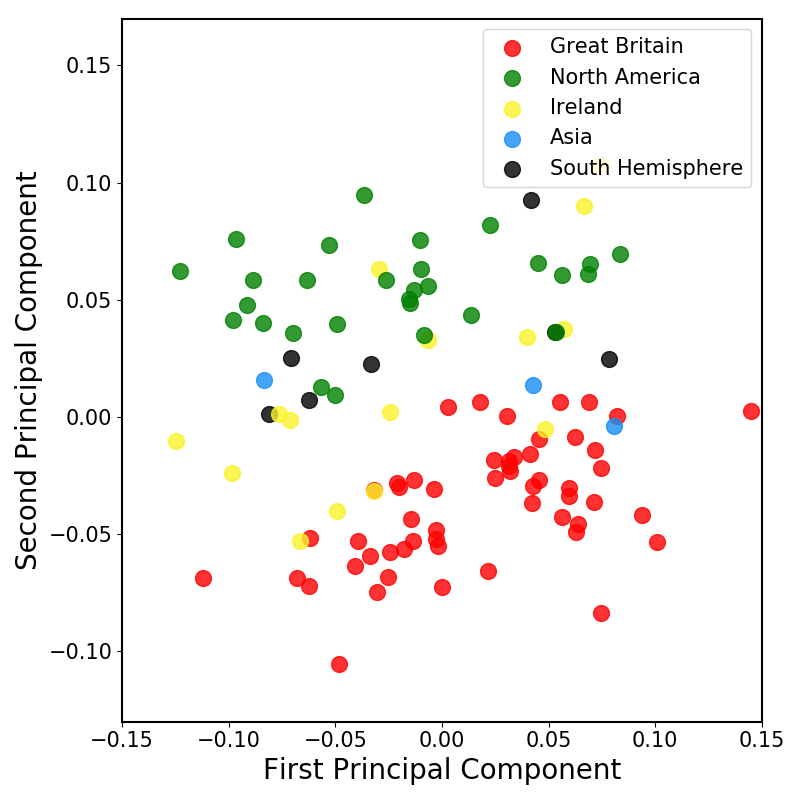} \vspace{0pt}
    \caption{Principal components of the learned speaker embeddings of multi-speaker ClariNet with 40-layer WaveNet, shown with genders and regions of the speakers.}
    \label{fig:pca}
\end{figure*}

\begin{table*}[t]
    \caption{Mean Opinion Score (MOS) ratings with 95\% confidence intervals, speaker classification accuracy, and speaker verification EERs (using 1 and 5 enrollment audios, denoted as EER-1 and EER-5, respectively) of multi-speaker ClariNet, Deep Voice 2, and Deep Voice 3 trained on the VCTK dataset. 
    We train the speaker classification and verification models using the VCTK and LibriSpeech datasets, respectively.
    The EER-1 and EER-5 on 50 held-out speakers from LibriSpeech are 6.30\% and 2.72\%, respectively. 
    }
    \centering
    \begin{tabular}{|c c|c|c|c|c|} \hline
    \multicolumn{2}{|c|}{\textbf{Multi-speaker Model}}         & \textbf{MOS}           & \textbf{Accuracy} & \textbf{EER-1}   & \textbf{EER-5} \\\hline
    \multicolumn{2}{|c|}{VCTK Ground Truth}                    &  ${\bf 4.26 \pm 0.38}$ & 100\%             & 12.8\%           & 5.4\%          \\\hline
    \multirow{3}{*}{Multi-speaker ClariNet} & 20-layer WaveNet & $3.75 \pm 0.42$        & 99.3\%            & 12.2\%           & 3.8\%          \\
                                            & 30-layer WaveNet & ${\bf 3.90 \pm 0.36}$  & 99.3\%            & 12.3\%           & 4.1\%          \\
                                            & 40-layer WaveNet & ${\bf 3.89 \pm 0.28}$  & 99.5\%            & 12.1\%           & 3.7\%          \\\hline
    \multirow{4}{*}{Deep Voice 2}           & 20-layer WaveNet & $3.19 \pm 0.44$        & 98.8\%            & 13.3\%           & 4.2\%          \\
                                            & 40-layer WaveNet & $3.36 \pm 0.37$        & 99.9\%            & 12.6\%           & 3.7\%          \\
                                            & 60-layer WaveNet & $3.50 \pm 0.23$        & 99.5\%            & 13.0\%           & 3.9\%          \\
                                            & 80-layer WaveNet & $3.63 \pm 0.31$        & 99.9\%            & 12.8\%           & 3.7\%          \\\hline
    Deep Voice 3                            & 30-layer WaveNet & $3.74 \pm 0.33$        & 99.9\%            & 12.8\%           & 4.0\%          \\\hline
    \end{tabular}
    \label{tab:clarinet_result}
\end{table*}

We compare the proposed multi-speaker ClariNet to Deep~Voice~2~(DV2) and Deep~Voice~3~(DV3).
Four DV2 models are evaluated with 20, 40, 60 and 80 layers WaveNet, respectively.\footnote{Independently trained WaveNet vocoder with softmax output.\label{Wavenet}}  
We also evaluate DV3 with 30-layer WaveNet vocoder~\textsuperscript{\ref{Wavenet}}. 

\textbf{Naturalness:} 
We randomly choose 10 speakers from the VCTK dataset and use a 10-sentence test set to synthesize a total of 100 audios for each model. 
To evaluate the quality of these synthesized audio, we run 5-scale mean opinion score (MOS) evaluations using the crowdMOS toolkit~\citep{ribeiro2011crowdmos} on Amazon Mechanical Turk.
The batches of samples from all these models were presented to raters, which naturally induces a comparison between the models.
We purposefully include ground truth samples in the set being evaluated, and the ground truth MOS provides a reference for calibrating the subjective evaluation. 

The results are shown in Table \ref{tab:clarinet_result}.
The results indicate that the quality of synthesized speech from multi-speaker ClariNet is higher than that of both DV2 and DV3 with WaveNet vocoder. The multi-speaker ClariNet with 20 layers (the fewest number of layers) outperforms DV2 with 80 layers (the largest number of layers), and the multi-speaker ClariNet with 30 layers outperforms DV3 with the same number of layers. 
From the results of DV2 and multi-speaker ClariNet, we also observe that the quality generally improves as we add more layers of the dilated convolution in WaveNet. 

\textbf{Speaker identity:} 
A multi-speaker TTS system that generates high fidelity and indistinguishable voices would result in high MOS, but fail to meet the desired objective of reproducing the various voices accurately. To show that our models can achieve high audio
quality synthesis while preserving the speaker identities, we synthesize a total of 2484 audios for each model using a 23-sentence test set on all 108 speakers from the VCTK dataset and use well-trained speaker classification and verification models for evaluations as suggested by \citep{dv2, arik2018voiceclone}. 

We follow the same setting to train a speaker classifier as in~\citep{dv2} on VCTK dataset to classify which of the 108 speakers a speech sample belongs to. 
The classification results in Table \ref{tab:clarinet_result} demonstrate that samples generated from multi-speaker ClariNet models are as distinguishable as the ground truth audios as well as those generated from the DV2 and DV3 counterparts.

Speaker verification is the task of authenticating the claimed identity of a speaker, based on a test audio and enrolled audios from the speaker. In particular, it performs binary classification to identify whether the test audio and enrolled audios are from the same speaker~\citep{Snyder2016}. We train the same speaker verification model as in~\citep{arik2018voiceclone} using the LibriSpeech dataset,  which consists of 820 hours of audio data from 2484 speakers. As a quantitative performance metric, the equal error-rate (EER) can be used to measure how close the audios generated from multi-speaker TTS models are to the ground truth audios. EERs are estimated by randomly pairing up utterances from the same or different speakers (50\% for each case) from the test audios generated by multi-speaker TTS models and the ground-truth samples from the VCTK dataset. We perform $40960$ trials to obtain the mean EERs using $1$ and $5$ enrollment audios, respectively. The results are listed in Table~\ref{tab:clarinet_result}, where the VCTK ground truth results represent EERs estimated from random pairing of utterances from the VCTK dataset. The EERs of multi-speaker ClariNet models are comparable to those of the VCTK ground truth, DV2, and DV3, which indicates that audios synthesized by multi-speaker ClariNet models are highly distinguishable. Note that, the EERs of the VCTK ground truth may be higher than those of the multi-speaker models, and this is caused by the discrepancy between the training LibriSpeech dataset and the testing VCTK dataset~\citep{arik2018voiceclone}.

\textbf{Speaker embedding:} In addition to evaluating the quality of the generated audio samples, we explore the latent space of the learned speaker embeddings by visualizing them in a lower dimensional space. The first two principal components of 32 dimensional speaker embeddings of multi-speaker ClariNet are plotted in Figure~\ref{fig:pca}. Each speaker is plotted as circle of a specific color depending on the gender or region of the speaker. Although the embeddings are randomly initialized and learned without any gender or region information of the speakers, the figure suggests that they encode discriminative information for our generative models. Gender of the speakers has such an apparent discriminative pattern that a linear classifier fit on the two dimensional space can classify the gender with high accuracy. We also observe apparent discriminative patterns for region of the speakers. In particular, Great Britain (red circle) and North America (green circle) regions can be perfectly separated by a linear classifier. Such apparent distinction among the gender and region groups also validates that our method can generate not only high quality but also distinguishable voices.





\section{Conclusion}
In this paper, we present multi-speaker ClariNet that can produce high-fidelity voices of various speakers. 
The shared speaker embeddings can learn discriminative features of each speaker since they serve as a bias to each component of the model.
Also, by directly conditioning on the hidden state from the bridge-net to the vocoder, the model can generate raw waveform from the text input without a separately trained vocoder. 
The results show that this end-to-end model with speaker embedding performs better than state-of-the-art multi-speaker models, even with fewer layers in the vocoder.

Despite the good performance of the model, we still think there exists a room for improvement. For instance, applying the parallel wave generation method described in \citep{Ping2018} to the current autoregressive vocoder could reduce huge amount of running time at inference.  
In addition, multi-speaker ClariNet can be used for voice cloning. One simple approach described in \citep{arik2018voiceclone} is training a speaker encoder model to predict the speaker embedding of an unseen speaker, which can then be plugged into the multi-speaker model. However, training a speaker encoder requires a larger dataset with more speakers than the VCTK dataset, since it is difficult to generalize speaker characteristics with only a hundred speakers. In the future, we would like to train our multi-speaker ClariNet model on the LibriTTS dataset~\citep{libritts}, which consists of 585 hours of audio recordings from 2456 speakers.


\renewcommand{\bibsection}{\section*{References}} {
\bibliographystyle{abbrvnat}
\bibliography{refs}}

\begin{thebibliography}{22}
\providecommand{\natexlab}[1]{#1}
\providecommand{\url}[1]{\texttt{#1}}
\expandafter\ifx\csname urlstyle\endcsname\relax
  \providecommand{\doi}[1]{doi: #1}\else
  \providecommand{\doi}{doi: \begingroup \urlstyle{rm}\Url}\fi

\bibitem[Arik et~al.(2017)Arik, Diamos, Gibiansky, Miller, Peng, Ping, Raiman,
  and Zhou]{dv2}
S.~Arik, G.~Diamos, A.~Gibiansky, J.~Miller, K.~Peng, W.~Ping, J.~Raiman, and
  Y.~Zhou.
\newblock Deep {V}oice 2: Multi-speaker neural text-to-speech.
\newblock In \emph{Advances in Neural Information Processing Systems}, pages
  2962--2970, 2017.

\bibitem[Ar{\i}k et~al.(2017)Ar{\i}k, Chrzanowski, Coates, Diamos, Gibiansky,
  Kang, Li, Miller, Ng, Raiman, Sengupta, and Shoeybi]{dv1}
S.~{\"O}. Ar{\i}k, M.~Chrzanowski, A.~Coates, G.~Diamos, A.~Gibiansky, Y.~Kang,
  X.~Li, J.~Miller, A.~Ng, J.~Raiman, S.~Sengupta, and M.~Shoeybi.
\newblock {Deep Voice}: Real-time neural text-to-speech.
\newblock In \emph{International Conference on Machine Learning}, 2017.

\bibitem[Arik et~al.(2018)Arik, Chen, Peng, Ping, and Zhou]{arik2018voiceclone}
S.~O. Arik, J.~Chen, K.~Peng, W.~Ping, and Y.~Zhou.
\newblock Neural voice cloning with a few samples.
\newblock In \emph{Advances in Neural Information Processing Systems}, pages
  10019--10029, 2018.

\bibitem[Fan et~al.(2015)Fan, Qian, Soong, and He]{fan2015multi}
Y.~Fan, Y.~Qian, F.~K. Soong, and L.~He.
\newblock Multi-speaker modeling and speaker adaptation for {DNN}-based {TTS}
  synthesis.
\newblock In \emph{{IEEE} International Conference on Acoustics, Speech and
  Signal Processing (ICASSP)}, pages 4475--4479, 2015.

\bibitem[Kingma and Ba(2015)]{Kingma2015}
D.~P. Kingma and J.~Ba.
\newblock Adam: {A} method for stochastic optimization.
\newblock In \emph{International Conference on Learning Representations}, 2015.

\bibitem[Ling et~al.(2015)Ling, Kang, Zen, Senior, Schuster, Qian, Meng, and
  Deng]{ling2015deep}
Z.-H. Ling, S.-Y. Kang, H.~Zen, A.~Senior, M.~Schuster, X.-J. Qian, H.~M. Meng,
  and L.~Deng.
\newblock Deep learning for acoustic modeling in parametric speech generation:
  A systematic review of existing techniques and future trends.
\newblock \emph{IEEE Signal Processing Magazine}, 2015.

\bibitem[Oord et~al.(2016)Oord, Dieleman, Zen, Simonyan, Vinyals, Graves,
  Kalchbrenner, Senior, and Kavukcuoglu]{wavenet}
A.~v.~d. Oord, S.~Dieleman, H.~Zen, K.~Simonyan, O.~Vinyals, A.~Graves,
  N.~Kalchbrenner, A.~Senior, and K.~Kavukcuoglu.
\newblock {WaveNet}: A generative model for raw audio.
\newblock \emph{arXiv preprint arXiv:1609.03499}, 2016.

\bibitem[Peng et~al.(2019)Peng, Ping, Song, and Zhao]{paranet}
K.~Peng, W.~Ping, Z.~Song, and K.~Zhao.
\newblock {Parallel Neural Text-to-Speech}.
\newblock \emph{arXiv preprint arXiv:1905.08459}, 2019.

\bibitem[Ping et~al.(2018)Ping, Peng, Gibiansky, Arik, Kannan, Narang, Raiman,
  and Miller]{dv3}
W.~Ping, K.~Peng, A.~Gibiansky, S.~O. Arik, A.~Kannan, S.~Narang, J.~Raiman,
  and J.~Miller.
\newblock Deep {V}oice 3: 2000-speaker neural text-to-speech.
\newblock In \emph{International Conference on Learning Representations}, 2018.

\bibitem[{Ping} et~al.(2019){Ping}, {Peng}, and {Chen}]{Ping2018}
W.~{Ping}, K.~{Peng}, and J.~{Chen}.
\newblock {ClariNet: Parallel Wave Generation in End-to-End Text-to-Speech}.
\newblock In \emph{International Conference on Learning Representations}, 2019.

\bibitem[Ribeiro et~al.(2011)Ribeiro, Flor{\^e}ncio, Zhang, and
  Seltzer]{ribeiro2011crowdmos}
F.~Ribeiro, D.~Flor{\^e}ncio, C.~Zhang, and M.~Seltzer.
\newblock Crowd{MOS}: An approach for crowdsourcing mean opinion score studies.
\newblock In \emph{{IEEE} International Conference on Acoustics, Speech and
  Signal Processing (ICASSP)}, 2011.

\bibitem[Shen et~al.(2018)Shen, Pang, Weiss, Schuster, Jaitly, Yang, Chen,
  Zhang, Wang, Ryan, Saurous, Agiomyrgiannakis, and Wu]{tacotron2}
J.~Shen, R.~Pang, R.~J. Weiss, M.~Schuster, N.~Jaitly, Z.~Yang, Z.~Chen,
  Y.~Zhang, Y.~Wang, R.~Ryan, R.~A. Saurous, Y.~Agiomyrgiannakis, and Y.~Wu.
\newblock Natural {TTS} synthesis by conditioning {WaveNet} on {M}el
  spectrogram predictions.
\newblock In \emph{{IEEE} International Conference on Acoustics, Speech and
  Signal Processing (ICASSP)}, pages 4779--4783, 2018.

\bibitem[Snyder et~al.(2016)Snyder, Ghahremani, Povey, Garcia-Romero, Carmiel,
  and Khudanpur]{Snyder2016}
D.~Snyder, P.~Ghahremani, D.~Povey, D.~Garcia-Romero, Y.~Carmiel, and
  S.~Khudanpur.
\newblock Deep neural network-based speaker embeddings for end-to-end speaker
  verification.
\newblock In \emph{IEEE Spoken Language Technology Workshop (SLT)}, pages
  165--170, 2016.

\bibitem[Sotelo et~al.(2017)Sotelo, Mehri, Kumar, Santos, Kastner, Courville,
  and Bengio]{char2wav}
J.~Sotelo, S.~Mehri, K.~Kumar, J.~F. Santos, K.~Kastner, A.~Courville, and
  Y.~Bengio.
\newblock {Char2Wav}: End-to-end speech synthesis.
\newblock In \emph{ICLR Workshop}, 2017.

\bibitem[Taigman et~al.(2018)Taigman, Wolf, Polyak, and Nachmani]{voiceloop}
Y.~Taigman, L.~Wolf, A.~Polyak, and E.~Nachmani.
\newblock {VoiceLoop}: Voice fitting and synthesis via a phonological loop.
\newblock In \emph{International Conference on Learning Representations}, 2018.

\bibitem[Wang et~al.(2017)Wang, Skerry{-}Ryan, Stanton, Wu, Weiss, Jaitly,
  Yang, Xiao, Chen, Bengio, Le, Agiomyrgiannakis, Clark, and Saurous]{tacotron}
Y.~Wang, R.~J. Skerry{-}Ryan, D.~Stanton, Y.~Wu, R.~J. Weiss, N.~Jaitly,
  Z.~Yang, Y.~Xiao, Z.~Chen, S.~Bengio, Q.~V. Le, Y.~Agiomyrgiannakis,
  R.~Clark, and R.~A. Saurous.
\newblock Tacotron: Towards end-to-end speech synthesis.
\newblock In \emph{Interspeech}, pages 4006--4010, 2017.

\bibitem[Wu et~al.(2015)Wu, Swietojanski, Veaux, Renals, and King]{wu2015study}
Z.~Wu, P.~Swietojanski, C.~Veaux, S.~Renals, and S.~King.
\newblock A study of speaker adaptation for {DNN}-based speech synthesis.
\newblock In \emph{Sixteenth Annual Conference of the International Speech
  Communication Association}, 2015.

\bibitem[Yamagishi et~al.(2009{\natexlab{a}})Yamagishi, Nose, Zen, Ling, Toda,
  Tokuda, King, and Renals]{vctk2009}
J.~Yamagishi, T.~Nose, H.~Zen, Z.~Ling, T.~Toda, K.~Tokuda, S.~King, and
  S.~Renals.
\newblock Robust speaker-adaptive hmm-based text-to-speech synthesis.
\newblock \emph{IEEE Transactions on Audio, Speech, and Language Processing},
  17\penalty0 (6):\penalty0 1208--1230, 2009{\natexlab{a}}.

\bibitem[Yamagishi et~al.(2009{\natexlab{b}})Yamagishi, Nose, Zen, Ling, Toda,
  Tokuda, King, and Renals]{yamagishi2009robust}
J.~Yamagishi, T.~Nose, H.~Zen, Z.-H. Ling, T.~Toda, K.~Tokuda, S.~King, and
  S.~Renals.
\newblock Robust speaker-adaptive {HMM}-based text-to-speech synthesis.
\newblock \emph{IEEE Transactions on Audio, Speech, and Language Processing},
  2009{\natexlab{b}}.

\bibitem[Yang et~al.(2016)Yang, Wu, and Xie]{yang2016training}
S.~Yang, Z.~Wu, and L.~Xie.
\newblock On the training of {DNN}-based average voice model for speech
  synthesis.
\newblock In \emph{{IEEE} Asia-Pacific Signal and Information Processing
  Association Annual Summit and Conference (APSIPA)}, pages 1--6, 2016.

\bibitem[Zen et~al.(2013)Zen, Senior, and Schuster]{ze2013statistical}
H.~Zen, A.~Senior, and M.~Schuster.
\newblock Statistical parametric speech synthesis using deep neural networks.
\newblock In \emph{2013 ieee international conference on acoustics, speech and
  signal processing}, pages 7962--7966. IEEE, 2013.

\bibitem[Zen et~al.(2019)Zen, Dang, Clark, Zhang, Weiss, Jia, Chen, and
  Wu]{libritts}
H.~Zen, V.~Dang, R.~Clark, Y.~Zhang, R.~J. Weiss, Y.~Jia, Z.~Chen, and Y.~Wu.
\newblock {LibriTTS}: {A} corpus derived from librispeech for text-to-speech.
\newblock \emph{arXiv preprint arXiv:1904.02882}, 2019.

\end{thebibliography}

\end{document}